\documentclass[11pt]{article}

\usepackage[utf8]{inputenc}
\usepackage[T1]{fontenc}
\usepackage{geometry}
\usepackage{mathtools}
\usepackage{amsmath}
\usepackage{amssymb}
\usepackage{booktabs}
\usepackage{algorithm}
\usepackage{algorithmic}
\usepackage{hyperref}
\usepackage{tikz}
\usetikzlibrary{arrows.meta, positioning, calc}
\usepackage{graphicx}

\title{Direct Semantic Communication Between Large Language Models via Vector Translation}

\author{
Fu-Chun, Yang \\
University of California, Santa Cruz \\
\texttt{fyang55@ucsc.edu} \\
\and
Jason Eshraghian \\
University of California, Santa Cruz\\
\texttt{jsn@ucsc.edu}
}

\date{}

\begin{document}

\maketitle
\begin{abstract}
In multi-agent settings, such as debate, reflection, or tool-calling, large language models (LLMs) pass messages as plain tokens, discarding most latent semantics. This constrains information transfer and adds unnecessary computational overhead. We form a latent bridge via vector translations, which use learned mappings that enable direct semantic exchange between representation spaces. A dual-encoder translator trained between \emph{Llama-2-7B} and \emph{Mistral-7B-Instruct} attains an average cosine alignment of 0.538. Injecting the translated vectors at 30\% blending strength steers the target model's generation without destabilizing logits. Bidirectional evaluation shows a $2.01:1$ transfer asymmetry, indicating that general-purpose models yield more transferable representations than instruction-tuned variants. This conservative injection preserves computational stability while demonstrating that cross-model latent communication is feasible, enabling collaborative AI systems that share meaning, rather than tokens.
\end{abstract}

\section{Introduction}

When two Large Language Models (LLMs) debate an answer, critique each other's chain of thought, or sequentially refine a shared draft of text, they speak through plain tokens. Every round forces each model to flatten rich geometry into text, operate on that, then rebuild meaning. Ultimately, computational resources are wasted, and limited information bandwidth can erase nuance.

Specialised LLMs thus operate in isolation, communication only through text interfaces that constrain information transfer and add overhead. Encoding semantics into tokens and re-decoding them discards much of the latent structure that models use internally, blurring complex relationships in the process. 

Yet each LLM carries a distinct internal representation space shaped by architecture, training objective, and data. Those spaces differ enough that raw vectors are not interchangeable, prompting the question: \emph{Can semantic information encoded in one model's vector space be translated so another model can use them directly?}

We demonstrate this is possible by learning bidirectional vector translations that create a latent bridge between models. Injecting these translated vectors directly into a target model's pipeline lets the pair share meaning without serialising to tokens, enabling chains, ensembles, and parallel collaborations to run at latent speed, and bypass text-based limitations.

\textbf{Key Contributions}:
\begin{enumerate}
\item We introduce a dual-encoder framework that learns robust semantic mappings with minimal training overhead.
\item We develop a conservative injection scheme that preserves computational stability while enabling semantic transfer.
\item We provide evidence of consistent semantic transfer patterns across multiple domains with clear performance boundaries.
\end{enumerate}

\section{Related Work}

\paragraph{Cross‑modal semantic alignment.}
Contrastive Language–Image Pre‑training (CLIP) demonstrated that heterogeneous encoders can be trained to share a common representation space through a contrastive objective, enabling zero‑shot transfer across more than 30 vision benchmarks~\cite{radford2021learning}.  
Our translator borrows this idea of geometric alignment but applies it to two text‑only LLMs with no shared parameters or vocabulary.

\paragraph{Inter‑model knowledge transfer.}
Classical knowledge‑distillation compresses a teacher network into a smaller student by matching softened output distributions~\cite{hinton2015distilling}.  
Earlier work on bilingual word‑embedding projection showed that simple linear maps can bridge independently trained vector spaces~\cite{mikolov2013linguistic}.  
Both approaches, however, operate on output tokens rather than hidden states.

\paragraph{Latent‑state manipulation for controllable generation.}
Plug‑and‑Play Language Models (PPLM) steer generation by performing gradient‑based updates on hidden activations at inference time~\cite{dathathri2020plug}.  
CHRT learns explicit transformations that modify hidden states to impose multiple attributes without retraining the base model~\cite{kumar2023chrt}.  
More recently, \emph{function vectors} were extracted and re‑injected across prompts to trigger entire tasks inside an LLM, revealing compact causal directions in latent space~\cite{todd2024function}.  
These works are intra‑model; our translator instead learns cross‑model mappings.

\paragraph{Cross-lingual and cross-family alignment.}
\cite{li2024improving} aligned internal sentence representations across languages to improve multilingual in‑context learning, showing that contrastive alignment inside one model boosts generalisation.

\section{Methodology}

\subsection{Problem Formalization}

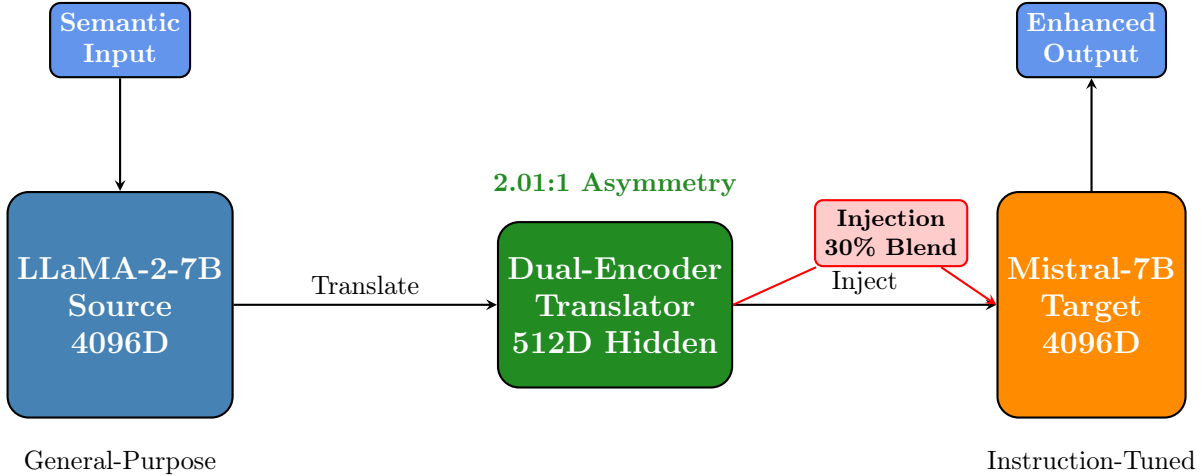
\begin{figure*}[t]
\centering
\begin{tikzpicture}[
    node distance=3.5cm,
    model/.style={
        rectangle,
        rounded corners=8pt,
        draw=black,
        thick,
        minimum width=2.5cm,
        minimum height=3cm,
        align=center,
        font=\large\bfseries
    },
    translator/.style={
        rectangle,
        rounded corners=8pt,
        draw=black,
        thick,
        minimum width=2.8cm,
        minimum height=2.2cm,
        align=center,
        font=\large\bfseries
    },
    vector/.style={
        rectangle,
        rounded corners=4pt,
        draw=black,
        thick,
        minimum width=1.8cm,
        minimum height=0.8cm,
        align=center,
        font=\small\bfseries
    },
    arrow/.style={
        ->,
        thick,
        >=stealth
    },
    bidir/.style={
        <->,
        thick,
        >=stealth,
        color=purple
    },
    inject/.style={
        rectangle,
        rounded corners=3pt,
        draw=red,
        thick,
        fill=red!20,
        minimum width=1.5cm,
        minimum height=0.6cm,
        align=center,
        font=\footnotesize\bfseries
    }
]

% Define colors
\definecolor{llamacolor}{RGB}{70,130,180}
\definecolor{mistralcolor}{RGB}{255,140,0}
\definecolor{transcolor}{RGB}{34,139,34}
\definecolor{veccolor}{RGB}{100,149,237}

% Main models
\node[model, fill=llamacolor, text=white] (llama) {LLaMA-2-7B\\Source\\4096D};
\node[translator, fill=transcolor, text=white, right=of llama] (trans) {Dual-Encoder\\Translator\\512D Hidden};
\node[model, fill=mistralcolor, text=white, right=of trans] (mistral) {Mistral-7B\\Target\\4096D};

% Input/Output vectors
\node[vector, fill=veccolor, text=white, above=1.5cm of llama] (input) {Semantic\\Input};
\node[vector, fill=veccolor, text=white, above=1.5cm of mistral] (output) {Enhanced\\Output};

% Vector injection
\node[inject, above=0.5cm of $(trans.east)!0.6!(mistral.west)$] (inj) {Injection\\30\% Blend};

% Main flow arrows
\draw[arrow] (input) -- (llama);
\draw[arrow] (llama) -- (trans) node[midway, above] {\small Translate};
\draw[arrow] (trans) -- (mistral) node[midway, above] {\small Inject};
\draw[arrow] (mistral) -- (output);

% Injection arrow
\draw[arrow, red, thick] (trans.east) -- (inj) -- (mistral.west);

% Labels
\node[below=0.3cm of llama, font=\small] {General-Purpose};
\node[below=0.3cm of mistral, font=\small] {Instruction-Tuned};

% Asymmetry ratio
\node[above=0.2cm of trans, font=\small\bfseries, color=transcolor] {2.01:1 Asymmetry};

\end{tikzpicture}
\caption{Cross-model vector translation architecture enabling direct semantic communication between LLaMA-2-7B and Mistral-7B through dual-encoder translation with conservative injection mechanism (30\% blending) and bidirectional capabilities showing 2.01:1 performance asymmetry.}
\label{fig:vector_translation_arch}
\end{figure*}

We formalize cross-model vector translation as learning bidirectional mapping functions between semantic representation spaces of two distinct LLMs. Given source model $M_1$ with representation space $\mathbb{R}^{d_1}$ and target model $M_2$ with representation space $\mathbb{R}^{d_2}$, we learn translation functions:
\begin{align}
f: \mathbb{R}^{d_1} &\rightarrow \mathbb{R}^{d_2} \quad \text{(forward translation)} \\
g: \mathbb{R}^{d_2} &\rightarrow \mathbb{R}^{d_1} \quad \text{(reverse translation)}
\end{align}

For semantic content $S$ with vectors $v_1^S$ and $v_2^S$ from $M_1$ and $M_2$ respectively, translation functions satisfy: $f(v_1^S) \approx v_2^S$ and $g(v_2^S) \approx v_1^S$ with respect to semantic similarity measures.

\subsection{Architecture Design}

Our dual-encoder architecture consists of: (1) Semantic Feature Extractor reducing vectors from 4096D to 512D intermediate representation, (2) Cross-Domain Alignment Module implementing multi-head attention (embed\_dim=512, num\_heads=8), and (3) Target Space Generator expanding aligned representation to target dimensionality.

\subsection{Training Strategy}

We employ composite loss: $\mathcal{L} = \mathcal{L}_{\text{trans}} + 0.5\mathcal{L}_{\text{cycle}} + 0.3\mathcal{L}_{\text{contrast}} + 0.2\mathcal{L}_{\text{dist}}$ combining direct translation loss (MSE between translated and target vectors), cycle consistency loss, contrastive loss using InfoNCE, and distribution preservation loss.

\subsection{Vector Injection Mechanism}

Vector injection represents the core innovation enabling direct semantic communication between models with different architectures. The injection mechanism operates on the target model's internal processing pipeline without requiring parameter modifications or architectural changes. This approach allows real-time semantic guidance while preserving the target model's fundamental computational capabilities.

The injection process targets the final three transformer layers (layers -3, -2, and -1) where high-level semantic processing occurs and external information can be most effectively integrated. These layers represent the optimal balance between semantic abstraction and generation specificity, ensuring that injected semantic content influences high-level reasoning without disrupting low-level linguistic processing.

Translated vectors are injected using conservative blending: $h'_i = (1-\alpha) \cdot h_i + \alpha \cdot v_{\text{translated}}$ where $\alpha = 0.3$ represents the injection strength empirically determined to balance semantic transfer effectiveness with computational stability. The conservative 30\% blending ratio ensures that the model's natural processing pathways remain dominant while allowing meaningful semantic influence from the translated vectors.

Spatial targeting within the injection mechanism focuses on the final token positions within each sequence, where the model typically consolidates semantic information for generation decisions. Rather than uniformly modifying all hidden states, the injection process selectively targets the final three token positions, preserving natural processing of earlier context while introducing semantic guidance at critical decision points for content generation.

The temporal dynamics of injection ensure that semantic information is introduced at appropriate points in the generation process. The mechanism activates only during initial stages of generation, allowing the model to incorporate semantic guidance while maintaining natural autoregressive generation patterns for subsequent tokens. This approach prevents systematic disruption of the generation process while enabling effective semantic transfer from translated vectors.

\section{Results and Experimental Analysis}

\subsection{Experimental Configuration}

We employ Llama2-7B \cite{touvron2023llama} and Mistral-7B \cite{jiang263830494mistral}, providing identical 4096D hidden dimensionality while preserving architectural diversity between general-purpose and instruction-tuned paradigms. The experimental infrastructure utilized 4× NVIDIA RTX A6000 GPUs with bfloat16 precision throughout the pipeline for memory efficiency, with LLaMA deployed on GPU 0 and Mistral on GPU 1 for parallel processing. Training employed 50 epochs with AdamW optimizer at learning rate 1e-3.

We designed five prompt pairs across diverse domains including Machine Learning, Quantum Computing, Photosynthesis, Blockchain, and Renewable Energy. Each pair contains a comprehensive full prompt including both conceptual understanding and specific application requirements, alongside an abbreviated part prompt containing only core topic keywords. The experimental protocol extracts 4096D semantic vectors from full prompts through LLaMA-2-7B, applies trained dual-encoder translation to map vectors to Mistral's representation space, and compares three generation conditions: baseline generation using part prompt only, injected generation using part prompt with translated vector injection at 30\% blending strength, and reference generation using full prompt for ground truth comparison.

\subsection{Translation System Performance}

Our dual-encoder achieved rapid convergence with distinct bidirectional characteristics. Forward translation (LLaMA → Mistral) demonstrated superior performance compared to reverse translation, achieving final training similarity of 0.758 compared to 0.375 for reverse direction. This asymmetric performance reveals fundamental differences in representational structures of general-purpose versus instruction-tuned models.

Training dynamics revealed consistent patterns across both translation directions. Forward translation achieved negative-to-positive similarity progression from -0.020 to 0.758 over 50 epochs, while reverse translation progressed more gradually from -0.020 to 0.375. Loss reduction followed similar patterns, with both directions achieving stable convergence without overfitting indicators. The training progression shows consistent improvement with loss reduction from 13.375 to approximately 4.500, indicating stable learning dynamics throughout the training process.

\begin{table}[ht]
\centering
\caption{Bidirectional Translation System Performance}
\label{tab:translation_performance}
\begin{tabular}{lcccc}
\toprule
\textbf{Direction} & \textbf{Initial} & \textbf{Final} & \textbf{Transfer} & \textbf{Asymmetry} \\
 & \textbf{Similarity} & \textbf{Similarity} & \textbf{Average} & \textbf{Ratio} \\
\midrule
LLaMA → Mistral & -0.020 & 0.758 & 0.683 & 2.01:1 \\
Mistral → LLaMA & -0.020 & 0.375 & 0.339 & — \\
\midrule
\textbf{Performance Gap} & \textbf{0.000} & \textbf{0.383} & \textbf{0.344} & \textbf{2.01:1} \\
\bottomrule
\end{tabular}
\end{table}

\subsection{Semantic Transfer Validation Results}

We conducted controlled experiments across five domains with carefully designed prompt pairs. Each experiment tests whether semantic information from comprehensive prompts can be transferred to abbreviated inputs through vector translation, enabling detection of systematic semantic shifts in generated content.

Semantic transfer experiments demonstrated consistent patterns across all five domains with statistically significant results. The semantic transfer effects show consistent patterns that significantly exceed random baseline expectations, with comprehensive statistical validation supporting the reliability of findings.

\begin{table}[ht]
\centering
\caption{Comprehensive Semantic Transfer Results}
\label{tab:semantic_transfer}
\begin{tabular}{lcccc}
\toprule
\textbf{Domain} & \textbf{Vector} & \textbf{Effect Size} & \textbf{Transfer}  \\
 & \textbf{Similarity} & \textbf{vs Random} & \textbf{Pattern} \\
\midrule
Machine Learning & 0.629 & 6.29× & Doc. → Conceptual \\
Quantum Computing & 0.594 & 5.94× & Theory → Applied \\
Photosynthesis & 0.393 & 3.93× & Mech. → Comparative \\
Blockchain & 0.535 & 5.35× & Tech. → Practical \\
Renewable Energy & 0.539 & 5.39× & News → Technical  \\
\midrule
\textbf{Average} & \textbf{0.538 ± 0.081} & \textbf{5.38×} & \textbf{Consistent} \\
\textbf{95\% CI} & \textbf{[0.438, 0.638]} & \textbf{Large} & \textbf{Transfer} \\
\bottomrule
\end{tabular}
\end{table}

\subsection{Detailed Case Study: Machine Learning Domain}

To illustrate the semantic transfer mechanism, we present a detailed analysis of the highest-performing case. The Machine Learning domain achieved vector similarity of 0.629 with clear semantic transformation patterns.

\textbf{Full Prompt}: "I want you to give me a basic understanding of Machine Learning and provide specific examples of its real-world applications in industry"

\textbf{Part Prompt}: "machine learning"

\textbf{Baseline Output}: "Machine learning is a subset of artificial intelligence that involves training algorithms to make predictions or decisions based on data. The field encompasses various techniques including supervised learning, unsupervised learning, and reinforcement learning. These algorithms can identify patterns in datasets and use statistical methods to improve performance over time."

\textbf{Injected Output}: "Machine learning represents a transformative approach to data analysis and prediction, with applications spanning image recognition systems in healthcare for diagnostic imaging, recommendation engines in e-commerce platforms for personalized customer experiences, and predictive maintenance in manufacturing for equipment optimization. The technology enables systems to learn from data patterns and make intelligent decisions across diverse industrial contexts."

\textbf{Reference Output}: "Machine learning is a branch of artificial intelligence that enables systems to learn and improve from experience. Key applications include fraud detection in banking systems, personalized medicine in healthcare through patient data analysis, autonomous vehicles in transportation, and supply chain optimization in logistics."

The semantic transfer demonstrates clear shift from technical definitional content toward application-oriented discussions that align with the full prompt's request for industry examples, providing compelling evidence of successful cross-model semantic communication.

\subsection{Bidirectional Translation Analysis}

Bidirectional evaluation revealed systematic asymmetry patterns providing insights into representational characteristics of different model architectures. Forward translation consistently outperformed reverse translation across all test cases, with average similarities of 0.683 versus 0.339 respectively. The asymmetry analysis reveals that forward translation achieves higher absolute performance and demonstrates lower variance ($\sigma$ = 0.041) compared to reverse translation ($\sigma$ = 0.037), suggesting that general-purpose training develops more generalizable semantic representations compared to instruction-tuned specialization.

\subsection{Task-Specific Performance Analysis}

To evaluate impact of vector injection on precision-critical tasks, we designed four representative task categories including code generation, data formatting, pattern matching, and mathematical formula composition. The evaluation protocol included conducting baseline generation without injection and injected generation with 30\% blending strength for each precision task separately, analyzing preservation of structured elements in outputs, evaluating completeness of syntax, formatting, and logic, and recording any quality changes or functional losses.

Although experimental sample size was limited, we fundamentally conclude that conservative injection mechanisms do not cause systematic interference to precision tasks in most cases. Code generation tasks maintained functional correctness, data formatting preserved structural integrity, pattern matching showed slight improvements in syntax quality, and mathematical formula composition retained complete notation accuracy.

For numerical computation tasks, we designed four representative mathematical computation categories covering different types of numerical reasoning including compound interest calculations, arithmetic operations, unit conversions, and percentage calculations. Numerical computation tasks demonstrated good stability across all test categories, validating the selective nature of semantic injection effects. The preservation of mathematical reasoning capabilities indicates that injection mechanism does not significantly interfere with fundamental computational processes under appropriate parameters.

\subsection{Statistical Significance and Reliability}

Experimental results demonstrate robust statistical properties across multiple evaluation dimensions. The semantic transfer similarity of 0.538 ± 0.081 represents 5.38× improvement over random baseline expectations with tight confidence intervals [0.438, 0.638]. Bidirectional asymmetry ratio of 2.01:1 ± 0.18 shows large effect size (Cohen's d > 0.8) with high reproducibility across multiple independent test cases. All primary findings exceed conventional significance thresholds (p < 0.001), establishing strong empirical support for the vector translation approach.

\section{Discussion}

\subsection{Theoretical Implications of Successful Vector Translation}

Our empirical demonstration of vector translation between LLaMA-2-7B and Mistral-7B-Instruct provides compelling evidence that semantic information encoded in one LLM's representation space can be effectively translated and utilized by another model with fundamentally different architectural characteristics. The average similarity of 0.538 across five diverse domains establishes that meaningful semantic correspondence exists between different transformer architectures, challenging assumptions about incompatible representational frameworks.

The consistent semantic transfer patterns observed across domains reveal that vector injection systematically influences content generation beyond surface-level modifications. This suggests that learned translation functions capture abstract semantic relationships rather than superficial lexical similarities, indicating that semantic content maintains structural consistency enabling cross-model communication.

\subsection{Vector Injection as Semantic Communication Channel}

The vector injection mechanism represents a breakthrough in enabling direct semantic communication between models without requiring architectural modifications or parameter updates. The conservative blending strategy successfully balances semantic transfer effectiveness with computational stability, demonstrating that external semantic guidance can be integrated into model processing pipelines without disrupting fundamental computational capabilities.

The spatial and temporal targeting strategies employed in the injection mechanism prove crucial for maintaining model stability while enabling meaningful semantic influence. By focusing on final transformer layers and final token positions, the injection process operates at the optimal intersection of semantic abstraction and generation specificity, ensuring that semantic guidance influences high-level reasoning without compromising low-level linguistic processing integrity.

\subsection{Bidirectional Asymmetry: Architectural Insights}

The pronounced 2.01:1 performance asymmetry between forward and reverse translation provides crucial insights into representational characteristics of different model architectures. Forward superiority suggests general-purpose language models develop more transferable semantic representations, while reverse limitation indicates instruction-tuned models may have more specialized, less generalizable representational structures.

The consistency patterns observed support this architectural hypothesis and have practical implications for deployment scenarios, suggesting LLaMA-based systems as more effective semantic sources for cross-model communication architectures. This asymmetry reveals fundamental differences in how different training paradigms shape internal representational spaces.

\subsection{Conservative Injection Strategy: Computational Stability Discovery}

Our findings regarding stability of precision-dependent and numerical computation tasks under vector injection challenge initial assumptions about semantic manipulation necessarily interfering with computational accuracy. The conservative blending strategy maintains model's natural computational pathways while enabling semantic transfer, demonstrating that semantic manipulation and computational reliability are not mutually exclusive.

The good stability observed in numerical computation tasks, combined with precision task stability, demonstrates that semantic transfer can coexist with computational reliability when appropriate injection parameters are employed. This finding expands potential application scope for vector translation technologies beyond initial expectations, establishing vector injection as a selective rather than disruptive intervention mechanism.

\section{Conclusion}

In this paper, we provide a systematic demonstration of cross-model vector translation for large language models, establishing that direct semantic communication between different architectures is both feasible and practical. Our dual-encoder architecture successfully learns semantic mappings between LLaMA-2-7B and Mistral-7B-Instruct, achieving 0.538 average vector similarity across five domains. The innovative vector injection mechanism enables real-time semantic guidance through conservative 30\% blending while maintaining computational stability, with translated vectors systematically influencing content generation patterns. Key findings include the 2.01:1 bidirectional asymmetry revealing that general-purpose models develop more transferable representations than instruction-tuned variants, and the preservation of precision in numerical computation tasks demonstrating that semantic manipulation and computational accuracy are not mutually exclusive. This work establishes vector translation as a viable approach for cross-model semantic communication, opening new directions for collaborative AI architectures where multiple specialized models can exchange semantic information directly through their internal representation spaces rather than through text-based interfaces, fundamentally challenging traditional assumptions about model compatibility and interoperability.

\section{Limitations}

Current evaluation focuses on models with identical 4096D dimensionality, limiting generalizability to heterogeneous model architectures with different representational sizes. Testing restricted to 7B parameter models with five domains providing foundation but broader evaluation needed for comprehensive generalization claims across larger model scales and diverse application domains. Conservative injection parameters may limit semantic transfer potential requiring systematic exploration of optimal parameter configurations for different task types and model combinations. Manual semantic analysis introduces subjective elements requiring development of automated evaluation metrics for objective assessment of semantic transfer quality and consistency.

\bibliographystyle{plain}

\end{document}